\documentclass[10pt,twocolumn,letterpaper]{article}

\usepackage{cvpr}              
\usepackage{tabularx}
\usepackage{booktabs,tabularx,array,xcolor}
\usepackage{graphicx}
\usepackage{cuted}
\usepackage{caption}
\definecolor{cvprblue}{rgb}{0.21,0.49,0.74}
\usepackage[pagebackref,breaklinks,colorlinks,allcolors=cvprblue]{hyperref}

\def\paperID{*****} 
\def\confName{CVPR}
\def\confYear{2026}

\title{Elastic-WAM: Adaptive Spatiotemporal Imagination for Efficient Robot Control}

\author{
    Haitian Li$^{1,2,*}$ \quad
    Tianfu Li$^{2,3,*}$ \quad
    Xuxi Zhao$^{2,4}$ \quad
    Zizheng Liu$^{2,5}$ \quad
    Yangcheng Yu$^{6}$ \quad \\
    Yiwen Hou$^{1,2}$ \quad
    Zeyu Xiao$^{1}$ \quad
    Mutian Xu$^{7}$ \quad
    Ziqian Yang$^{2,8}$ \quad
    Haonan Chen$^{1}$ \quad
    Lin Shao$^{1,2,\dagger}$
    \\[0.5em]
    $^{1}$National University of Singapore \quad $^{2}$RoboScience \\
    $^{*}$Equal contribution \quad
    $^{\dagger}$Corresponding author
}

\begin{document}
\maketitle

\begin{strip}
    \centering
    \includegraphics[width=\linewidth]{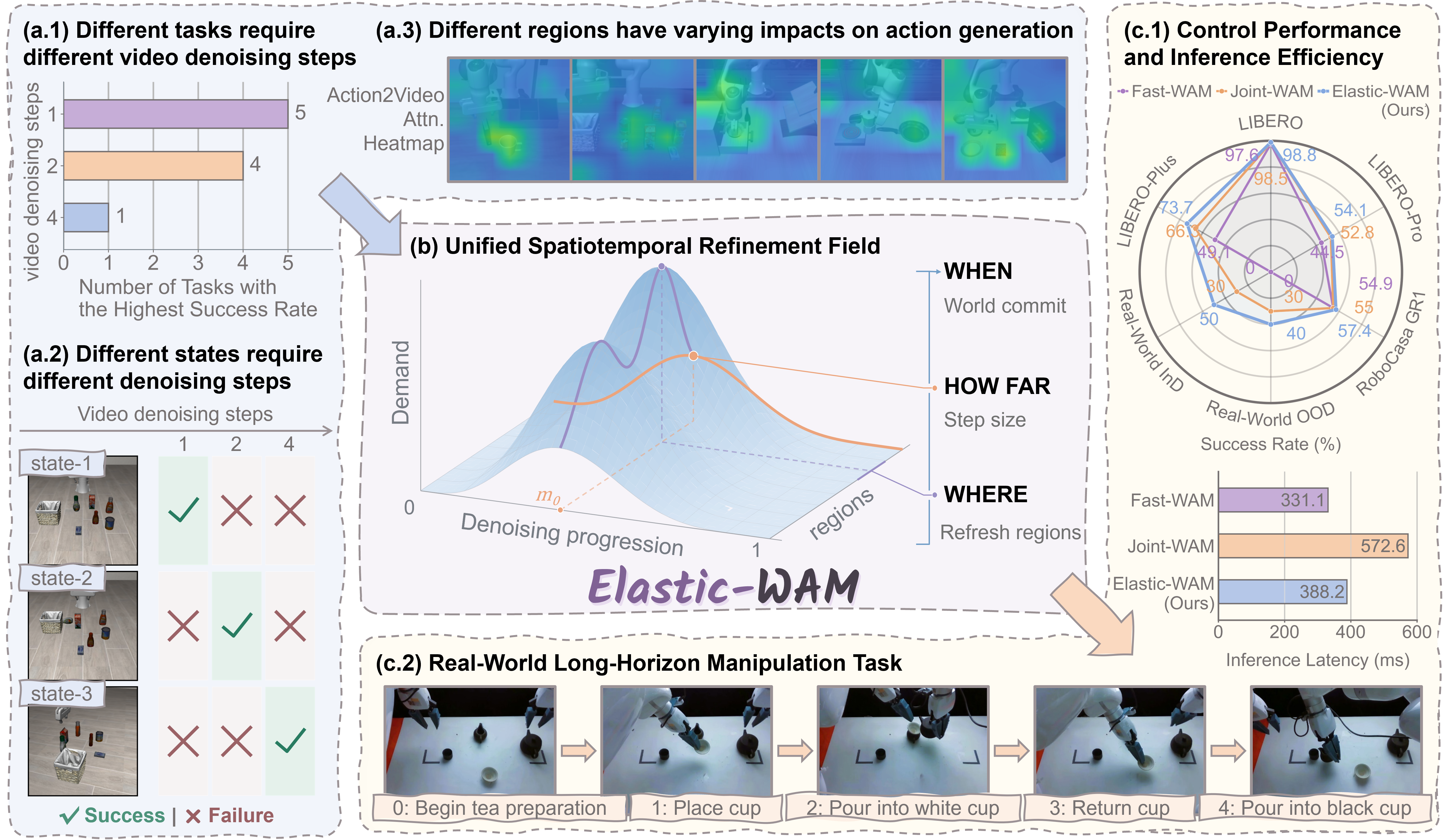}
    \captionsetup{hypcap=false}
    \captionof{figure}{
        \textbf{Overview of Elastic-WAM.}
        (a) Empirical observations reveal heterogeneous temporal refinement requirements across tasks and states, as well as varying spatial relevance to action generation.
        (b) So our unified spatiotemporal refinement field determines WHEN to stop world refinement, HOW FAR to advance denoising, and WHERE to refresh future video regions.
        (c) As a result, Elastic-WAM improves control performance and inference efficiency across simulation benchmarks and real-world long-horizon manipulation.
    }
    \label{fig:teaser}
    \captionsetup{hypcap=true}
\end{strip}

\begingroup
\renewcommand{\thefootnote}{}
\footnotetext[0]{\textbf{Affiliations 3--8:}
$^{3}$The Hong Kong University of Science and Technology (Guangzhou);
$^{4}$Zhejiang University;
$^{5}$Harbin Institute of Technology (Shenzhen);
$^{6}$Tsinghua University;
$^{7}$Nanyang Technological University;
$^{8}$Xi'an Jiaotong-Liverpool University.}
\endgroup
\begin{abstract}
World-Action Models improve robot control by predicting future visual dynamics, but iterative video denoising incurs substantial inference overhead.
Recent approaches reduce this cost through partial future denoising, yet typically rely on fixed refinement budgets.
Our empirical observations reveal that preferred refinement depths vary across tasks and states, while different future regions exhibit varying relevance to action generation.
Motivated by these findings, we propose \textbf{Elastic-WAM}, which adaptively allocates world refinement computation across temporal denoising time and spatial future space.
A unified spatiotemporal refinement demand field governs when to stop world refinement, how far to advance each denoising step, and which regions to refresh.
We parameterize this field with a shared temporal basis and spatial amplitudes, and learn it through a three-stage action-grounded supervised procedure.
At inference, the learned field guides recurrent sparse world refinement to improve task success while reducing inference overhead.
Evaluations across diverse simulation benchmarks and real-world long-horizon manipulation demonstrate state-of-the-art control performance with reduced inference latency.
\textbf{Project page:} \url{https://lihaitian.com/Elastic-WAM}
\end{abstract}    
\section{Introduction}
\label{sec:intro}

World-Action Models (WAMs) have shown that predicting future visual dynamics can improve robot control~\citep{ye2026world,bi2026motus,agarwal2026cosmos,team2026motubrain,chen2026objectworld}.
However, iterative denoising of future videos incurs substantial inference overhead~\citep{akbari2026flash,li2026slip}.
Recent studies show that intermediate video representations can already support action prediction without complete denoising~\citep{pai2025mimic,ma2026dit4dit,li2026causal}.
Explicit video generation can even be removed at inference while retaining future prediction as a training objective~\citep{yuan2026fast}.
Nevertheless, these approaches typically adopt a fixed future refinement budget, regardless of the current task or execution state~\citep{guo2026unified,chen2026mvg}.
Consequently, a fixed budget may waste computation in some states while limiting action quality in others~\citep{sun2026sants}.

To investigate this limitation, we examine how future video refinement affects downstream action quality across different tasks and states on 10 selected tasks from LIBERO-Plus benchmark~\citep{fei25libero-plus}.
As demonstrated in Fig.~\ref{fig:teaser} (a.1), the number of denoising steps yielding the highest success rate varies from one to four.
Even within the same task, different states favor different refinement depths, as shown in Fig.~\ref{fig:teaser} (a.2).
Beyond refinement depth, action-to-video attention maps reveal uneven attention across future video regions (shown in Fig.~\ref{fig:teaser} (a.3)), motivating selective refinement of regions most relevant to action generation~\citep{yu2026maskwam}.
Together, these observations motivate adaptive allocation of world refinement computation across both temporal denoising time and spatial future video regions.

To this end, we propose \textbf{Elastic-WAM}, which models the demand for world refinement through \textbf{a unified spatiotemporal field}, as illustrated in Fig.~\ref{fig:teaser} (b).
Its total mass governs when to stop world refinement, its temporal profile determines how far to advance each denoising step, and its spatial variation identifies which future video regions to refresh.
We parameterize this field with a shared temporal basis and spatial amplitudes that separate global refinement demand from relative regional priorities.
We learn this field through three stages of action-grounded supervision, progressively optimizing the temporal profile, relative spatial priorities, and overall refinement demand.
At inference, we start with a dense joint denoising step, use the learned field to guide subsequent sparse world refinement, and continue action denoising after world refinement stops.

We evaluate Elastic-WAM across in-distribution~\citep{liu2023libero}, out-of-distribution~\citep{fei25libero-plus,zhou2025libero}, and more complex manipulation simulation benchmarks~\citep{bjorck2025gr00t}, as well as a real-world long-horizon tea preparation task.
Our results demonstrate state-of-the-art control performance among methods without large-scale embodied pretraining, while reducing inference latency by 32.2\% compared with Joint-WAM on RoboCasa-GR1, as shown in Fig.~\ref{fig:teaser}(c).

Our contributions are summarized as follows:
\begin{itemize}
    \item We formulate adaptive world refinement through a unified spatiotemporal demand field, jointly governing \textbf{when} to stop, \textbf{how far} to advance, and \textbf{where} to refresh future video representations.
    \item We develop an action-grounded supervised learning framework and recurrent sparse inference strategy, enabling adaptive refinement with a frozen WAM backbone.
    \item We demonstrate state-of-the-art performance across simulation benchmarks and real-world manipulation tasks, while substantially reducing inference latency.
\end{itemize}
\section{Related Work}
\label{sec:related_work}

\noindent\textbf{World-Action Models.}
Vision-Language-Action (VLA) models have evolved from directly predicting robot actions~\citep{brohan2023rt,kim2024openvla,black2024pi_0,intelligence2025pi_05,he2026elan4d,xie2026dynamicvla} to leveraging future visual predictions to guide action generation~\citep{zhao2025cot,cen2025worldvla,zhang2026dreamvla,wang2026lamp,wang2026openwam}.
World-Action Models (WAMs) further extend the VLA paradigm by jointly modeling future visual dynamics and robot actions~\citep{ye2026world,bi2026motus,kim2026cosmos,team2026motubrain}.
To improve inference efficiency, recent approaches exploit partially denoised video representations~\citep{pai2025mimic,ma2026dit4dit,li2026causal}, modality-aware step distillation~\citep{akbari2026flash}, or compact video experts with sparse latents and asymmetric denoising~\citep{li2026efficient}.
Other methods eliminate explicit future generation at inference~\citep{yuan2026fast} or preserve future conditioning through sparse video--action interactions~\citep{zhao2026faster}.
Complementary to these approaches, our Elastic-WAM adaptively allocates computation within the recurrent world refinement process of a frozen WAM, across both temporal denoising time and spatial future video regions.

\begin{figure*}[t]
    \centering
    \includegraphics[width=\textwidth]{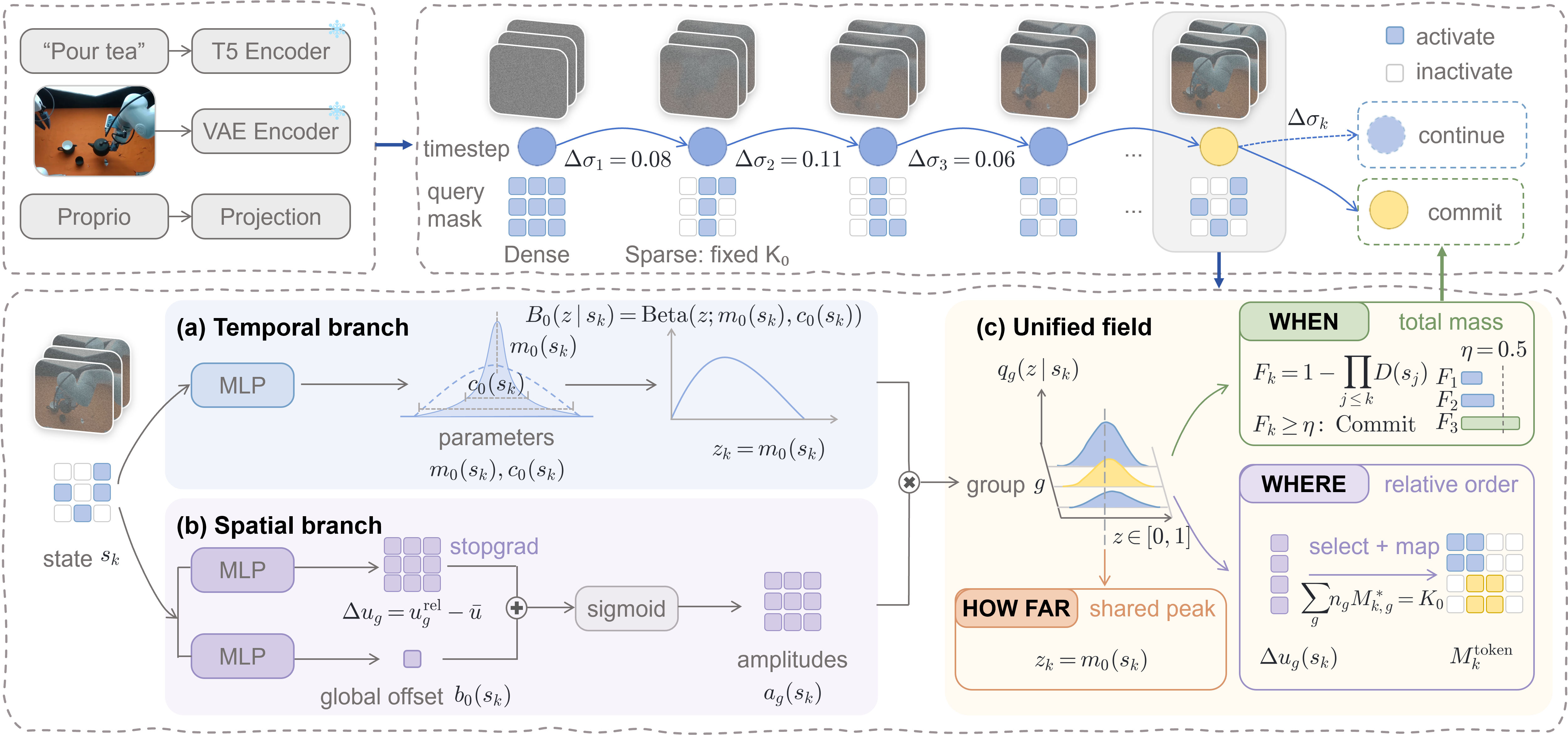}
    \caption{
        \textbf{Pipeline of Elastic-WAM.}
        \textbf{Top:} The inference pipeline starts with dense joint refinement, followed by adaptive sparse world refinement with state-dependent denoising progression and spatial token selection.
        Once the world is committed, action refinement continues using the cached world representations.
        \textbf{Bottom:} The unified spatiotemporal refinement field consists of (a) a temporal branch that predicts a shared Beta distribution over denoising progression, (b) a spatial branch that estimates group-wise refinement amplitudes, and (c) a unified field that determines WHEN to commit the world, HOW FAR to advance denoising, and WHERE to refresh future.
    }
    \label{fig:pipeline}
\end{figure*}

\noindent\textbf{Temporally Adaptive Denoising.}
Diffusion sampling has been accelerated through implicit samplers, efficient ODE solvers, and consistency distillation~\citep{song2020denoising,lu2022dpm,luo2023latent}, typically using predefined schedules across inputs.
Recent approaches further adapt denoising computation to individual inputs: AdaDiff learns sample-specific step budgets~\citep{zhang2025adadiff}, while AdaptiveDiffusion skips redundant noise predictions based on intermediate latent changes~\citep{ye2024training}.
For robotic control, D3P dynamically allocates action denoising steps according to state-dependent action importance~\citep{yu2025d3p}, while SANTS learns adaptive stopping and noise progression for WAMs based on downstream action quality~\citep{sun2026sants}.
While these methods adapt computation along the denoising trajectory, Elastic-WAM jointly governs temporal refinement and spatial future token updates through a unified spatiotemporal refinement field.

\noindent\textbf{Spatially Adaptive Computation.}
Spatial redundancy in diffusion transformers has motivated methods that concentrate computation on selected tokens or attention interactions.
VSA introduces trainable sparse attention by identifying important token interactions~\citep{zhang2026faster}, while VORTA dynamically routes computation among sparse attention patterns~\citep{sun2026vorta}.
Jenga combines dynamic attention carving with progressive resolution generation~\citep{zhang2026training}.
Beyond attention sparsification, DyDiT dynamically adjusts computation across denoising timesteps and spatial locations~\citep{zhao2025dynamic}, while token-wise feature caching selectively reuses intermediate features across denoising steps~\citep{zou2024accelerating}.
Elastic-WAM learns to selectively refresh future video queries based on their contribution to downstream action prediction, preserving full logical attention context and coordinating spatial refinement with temporal decisions through a unified field.
\section{Methodology}
\label{sec:method}


As illustrated in Fig.~\ref{fig:pipeline}, Elastic-WAM allocates world refinement computation according to the remaining refinement demand at each state.
We model this demand over denoising progression and future video space through a unified spatiotemporal field, where geometry-aware local curves describe the refinement needs of different regions (Sec.~\ref{sec:refinement-field}).
Since future video regions share the same denoising trajectory, we use a common temporal profile while allowing each region to have a different refinement amplitude.
Together, these components determine when to stop world refinement, how far to advance each denoising step, and which future video regions to refresh (Sec.~\ref{sec:conditional}).

With the pretrained WAM frozen, we learn the field through three stages of action-grounded supervision: first its temporal profile, then the relative refinement priorities of future video regions, and finally the overall demand for continued world refinement (Sec.~\ref{sec:training}).
At inference, Elastic-WAM begins with a dense joint denoising step, recurrently refreshes selected future video queries while preserving the full logical attention context, and continues action denoising using the cached video representation after world refinement stops (Sec.~\ref{sec:inference}).

\subsection{Unified Spatiotemporal Refinement Field}
\label{sec:refinement-field}
We unify the temporal and spatial refinement demand within a single spatiotemporal field.
To model spatially varying refinement demand, we partition the $N_F=196$ future video tokens into $G=64$ geometry-aware groups that preserve temporal and camera-view boundaries.
Each group $g$ contains $n_g\in\{1,2,4\}$ tokens, with $\sum_{g=1}^{G}n_g=N_F$.
For each group, we define a local refinement curve that describes how its remaining demand is distributed along the normalized denoising progression $z\in[0,1]$.
\begin{equation}
q_g(z\mid s)
=
a_g(s)\,
\operatorname{Beta}\!\left(z;m_g(s),c_g(s)\right),
\label{eq:local-refinement-curve}
\end{equation}
Here, $a_g(s)$ represents the remaining refinement demand of group $g$, while $\operatorname{Beta}(\cdot;m_g(s),c_g(s))$ denotes a Beta density parameterized by its mode $m_g(s)$ and concentration $c_g(s)$.

Since the Beta density is normalized, the total mass of each local refinement curve equals its amplitude $a_g(s)$.
We aggregate these local curves according to their respective token counts to obtain the global refinement demand:
\begin{equation}
\begin{aligned}
\int_0^1 q_g(z\mid s)\,dz &= a_g(s),\\
Q(z\mid s) &= \sum_{g=1}^{G}
\frac{n_g}{N_F}\,q_g(z\mid s).
\end{aligned}
\label{eq:global-refinement-field}
\end{equation}
Together, the local curves $\{q_g(z\mid s)\}_{g=1}^{G}$ form the unified spatiotemporal field, while $Q(z\mid s)$ represents its spatially aggregated refinement demand.
The total mass of $Q(z\mid s)$ captures the overall remaining refinement demand, its distribution along $z$ characterizes preferred denoising progression, and the local amplitudes $a_g(s)$ indicate how demand varies across future video regions.

\subsection{Refinement Field Parameterization}
\label{sec:conditional}
While the general formulation allows each region to have an independent temporal profile, all future video tokens follow a shared denoising trajectory in the underlying WAM.
We therefore parameterize each local refinement curve using a shared temporal basis $B_0(z\mid s)$ and a region-specific amplitude $a_g(s)$.
\begin{equation}
\begin{aligned}
q_g(z\mid s) &= a_g(s)B_0(z\mid s),\\
B_0(z\mid s) &= \operatorname{Beta}\!\left(z;m_0(s),c_0(s)\right).
\end{aligned}
\label{eq:shared-temporal-basis}
\end{equation}

Substituting this parameterization into the global aggregation yields
\begin{equation}
\begin{aligned}
Q(z\mid s) &= D(s)B_0(z\mid s),\\
D(s) &= \sum_{g=1}^{G}\frac{n_g}{N_F}a_g(s),
\end{aligned}
\label{eq:factorized-refinement-field}
\end{equation}
Since $B_0(z\mid s)$ is normalized, $D(s)$ represents the total remaining refinement demand, while $B_0(z\mid s)$ describes its distribution along the progression of denoising.

To separate overall refinement demand from relative spatial priorities, we center the region-wise scores by their token-weighted mean:
\begin{equation}
\begin{aligned}
\Delta u_g(s)
&= u_g^{\mathrm{rel}}(s)
- \sum_{j=1}^{G}\frac{n_j}{N_F}u_j^{\mathrm{rel}}(s),\\
\sum_{g=1}^{G}\frac{n_g}{N_F}\Delta u_g(s) &= 0.
\end{aligned}
\label{eq:relative-amplitude}
\end{equation}

We then combine the relative spatial scores with a shared offset $b_0(s)$ to define each group's refinement amplitude:
\begin{equation}
a_g(s)
=
\operatorname{sigmoid}\!\left(
b_0(s)+\operatorname{stopgrad}\!\left[\Delta u_g(s)\right]
\right).
\label{eq:absolute-relative-amplitude}
\end{equation}
The shared offset $b_0(s)$ controls the overall refinement demand $D(s)$, while $\Delta u_g(s)$ determines the relative priority of each region.
The stop-gradient operation prevents the learning of absolute demand from modifying the relative spatial scores.

\noindent\textbf{WHEN.}
We interpret the global refinement demand $D(s_k)$ as the conditional continuation probability and construct the cumulative \textit{World Commit} probability across refinement steps:
\begin{equation}
\begin{aligned}
C_k &= D(s_k),\\
\Delta H_k &= -\log\!\left(\operatorname{clamp}(C_k,\varepsilon,1)\right),\\
F_k &= 1-\exp\!\left(-\sum_{j\leq k}\Delta H_j\right)\\
&= 1-\prod_{j\leq k}C_j.
\end{aligned}
\label{eq:curve-native-when}
\end{equation}
We trigger \textit{World Commit} when $F_k\geq\eta$, with $\eta=0.5$.

\noindent\textbf{HOW FAR.}
Given continuation, we normalize the global refinement field to obtain the temporal progression distribution and use its mode to determine the denoising step size:
\begin{equation}
\begin{aligned}
p(z\mid s_k,\mathrm{CONTINUE})
&= \frac{Q(z\mid s_k)}{D(s_k)}
= B_0(z\mid s_k),\\
z_k &= m_0(s_k),\\
\Delta\sigma_k
&= \delta_{\min}
+ z_k(\delta_{\max}-\delta_{\min}).
\end{aligned}
\label{eq:conditional-progression}
\end{equation}
We set $\delta_{\min}=0.05$ and $\delta_{\max}=\min(0.20,\sigma_k-\sigma_{\mathrm{floor}})$ to constrain the denoising progression within the admissible noise interval.

\noindent\textbf{WHERE.}
Since the shared temporal basis and global offset do not affect spatial ordering, we select future video regions according to their relative refinement demand under an exact-token budget $K_0$:
\begin{equation}
\begin{aligned}
M_k^{*}
&= \arg\max_{M\in\{0,1\}^{G}}
\sum_{g=1}^{G} n_g M_g \Delta u_g(s_k),\\
\mathrm{s.t.}\quad
&\sum_{g=1}^{G} n_g M_g=K_0.
\end{aligned}
\label{eq:exact-k-where}
\end{equation}
We use an externally specified budget of $K_0=96$ tokens per sparse update by default, while the learned spatial scores determine which regions to refresh.

\subsection{Action-Grounded Refinement Field Learning}
\label{sec:training}

With the pretrained WAM frozen, we learn the refinement field in three stages, successively optimizing the temporal parameters $(m_0,c_0)$, the relative spatial scores $u_g^{\mathrm{rel}}$, and the global offset $b_0$.

\noindent\textbf{Temporal Profile (HOW FAR).}
To learn the temporal profile, we compare immediate \textit{World Commit} with continuations of different denoising distances from the same recurrent state $s$.
Each continuation performs one additional joint denoising step, followed by a forced \textit{World Commit} and complete \textit{Action Finish}.
We evaluate the resulting action sequence $y$ using a terminal action error:
\begin{equation}
E(s,y)
=
0.6E_{\mathrm{seq}}(s,y)
+
0.4E_{\Delta}(s,y),
\label{eq:terminal-action-error}
\end{equation}
where $E_{\mathrm{seq}}$ measures action-sequence error and $E_{\Delta}$ measures the error in consecutive action differences.
We convert the candidate errors into a soft target over denoising distances, assigning higher probabilities to candidates with lower terminal errors, to supervise the temporal distribution $B_0(z\mid s)$.

The same comparisons also provide stopping supervision by evaluating immediate \textit{World Commit} against the expected error of continuing under the learned temporal policy:
\begin{equation}
\begin{aligned}
E_{\mathrm{continue}}^{\pi}(s)
&= \sum_{\delta} P_{\pi}(\delta\mid s)E_{\delta}(s),\\
A_{\mathrm{stop}}(s)
&= E_{\mathrm{continue}}^{\pi}(s)-E_{\mathrm{commit}}(s),\\
p_{\mathrm{stop}}^{*}(s)
&= \operatorname{sigmoid}\!\left(
\frac{A_{\mathrm{stop}}(s)}{T(s)}
\right).
\end{aligned}
\label{eq:temporal-teacher}
\end{equation}
Here, $E_{\delta}(s)$ and $E_{\mathrm{commit}}(s)$ denote the terminal action errors of continuation by $\delta$ and immediate \textit{World Commit}, respectively, while $T(s)$ controls the softness of the target.
The resulting stopping supervision is used to train a temporal teacher, whose frozen hazard predictions guide the learning of overall refinement demand in the final stage.

\noindent\textbf{Spatial Priorities (WHERE).}
With the temporal profile fixed, we learn the relative spatial scores $u_g^{\mathrm{rel}}(s)$ by comparing dense and sparse future video updates at the same denoising state.
Starting from a dense joint update at noise level $\sigma_0$, we obtain the recurrent context and use the frozen temporal policy to select the next noise level $\sigma_1$.
At $\sigma_1$, we compare a no-gradient dense teacher with a differentiable sparse student using the same teacher-forced state $x_{\mathrm{cur}}$.
To optimize the discrete exact-budget mask, we use a straight-through relaxation that preserves hard selection in the forward pass while allowing gradients to flow through its soft counterpart:
\begin{equation}
\begin{aligned}
M^{\mathrm{ST}}
&= M^{\mathrm{hard}} + M^{\mathrm{soft}}
- \operatorname{stopgrad}(M^{\mathrm{soft}}),\\
\mathcal{L}_{\mathrm{where}}
&= \mathcal{L}_{\mathrm{action\text{-}GT}}
+ \mathcal{L}_{\mathrm{action\text{-}Dense}}\\
&\quad + 0.1\mathcal{L}_{\mathrm{video\text{-}GT}}
+ 0.01\mathcal{L}_{\mathrm{budget}}.
\end{aligned}
\label{eq:where-learning}
\end{equation}
The two action losses supervise the sparse student's predicted action flow against ground-truth flow and the dense teacher, respectively, while the video and budget terms provide auxiliary flow supervision and soft-budget regularization.
The WAM remains frozen while gradients propagate through its computations to optimize the spatial scores.
After training, we freeze the relative spatial branch to preserve the learned region priorities.

\noindent\textbf{Overall Demand (WHEN).}
With the temporal profile and relative spatial scores fixed, we learn the global offset $b_0(s)$ to control the overall refinement demand $D(s)$.
We use hazard predictions from the frozen temporal teacher to construct target continuation masses $C_k^*$, which are mapped to the corresponding global offsets $b_0^*(s_k)$ through the aggregated regional amplitudes:
\begin{equation}
\begin{aligned}
C_k^{*}
&= \exp\!\left(-\Delta H_k^{\mathrm{teacher}}\right),\\
C_k^{*}
&= \sum_{g=1}^{G}\frac{n_g}{N_F}
\operatorname{sigmoid}\!\left(
b_0^{*}(s_k)+\Delta u_g(s_k)
\right).
\end{aligned}
\label{eq:absolute-mass-target}
\end{equation}
Since the aggregated regional amplitudes increase monotonically with $b_0^*(s_k)$, this constraint uniquely determines the target global offset.

Since \textit{World Commit} depends on accumulated continuation probabilities, we train $b_0$ using complete recurrent prefixes rather than isolated states.
We optimize the following objective to supervise cumulative stopping behavior while retaining a terminal action-quality correction:
\begin{equation}
\begin{aligned}
\mathcal{L}_{\mathrm{seq}}
&= \mathcal{L}_{b_0}
+ \mathcal{L}_{\mathrm{conditional}}
+ \mathcal{L}_{\mathrm{cumulative}}\\
&\quad + 0.5\mathcal{L}_{\mathrm{commit\text{-}time}},\\
\mathcal{L}_{\mathrm{final}}
&= \mathcal{L}_{\mathrm{seq}}
+ 0.1\mathcal{L}_{\mathrm{terminal\text{-}CF}}.
\end{aligned}
\label{eq:absolute-demand-learning}
\end{equation}
Here, $\mathcal{L}_{b_0}$ supervises the target global offset, $\mathcal{L}_{\mathrm{conditional}}$ aligns the conditional continuation mass, and $\mathcal{L}_{\mathrm{cumulative}}$ and $\mathcal{L}_{\mathrm{commit\text{-}time}}$ constrain cumulative stopping probabilities and commit timing across recurrent prefixes.
Finally, $\mathcal{L}_{\mathrm{terminal\text{-}CF}}$ provides a low-weight correction based on terminal action quality at newly visited states, while the temporal profile and relative spatial branch remain frozen.

\subsection{Refinement Field-Guided Sparse Inference}
\label{sec:inference}

At each replanning cycle, Elastic-WAM begins with a dense joint denoising step over all future video tokens, establishing the recurrent state and layer-wise key--value cache.
The learned refinement field then determines whether to continue world refinement and, if so, selects the next denoising progression and the future video regions to refresh (Sec.~\ref{sec:conditional}).
All video and action latents are advanced to the next noise level, while the selected exact-token mask determines which future video queries are recomputed in the subsequent step.
This process repeats with updated refinement decisions at each step.


\begin{table*}[t]
    \centering
    \begin{minipage}[t]{0.485\textwidth}
        \centering
        \caption{LIBERO success rates (\%). The best and second-best results are highlighted in \textbf{bold} and \underline{underlined}, respectively.}
        \label{tab:libero}
        \vspace{0.3em}
        \begingroup
        \footnotesize
        \renewcommand{\arraystretch}{1.17}
        \setlength{\tabcolsep}{1pt}
        \setlength{\arrayrulewidth}{0.35pt}
        \begin{tabularx}{\linewidth}{@{}>{\raggedright\arraybackslash}p{0.39\linewidth}*{4}{>{\centering\arraybackslash}X}|>{\centering\arraybackslash}X@{}}
            \toprule
            Method & Spat. & Obj. & Goal & Long & Avg. \\
            \midrule
            $\pi_{0.5}$~\citep{intelligence2025pi_05} & 98.8 & 98.2 & \underline{98.0} & 92.4 & 96.9 \\
            LingBot-VA~\citep{li2026causal} & 98.5 & 99.6 & 97.2 & \textbf{98.5} & \underline{98.5} \\
            Motus~\citep{bi2026motus} & 96.8 & \underline{99.8} & 96.6 & 97.6 & 97.7 \\
            Fast-WAM~\citep{yuan2026fast} & 98.2 & \textbf{100.0} & 97.0 & 95.2 & 97.6 \\
            Joint-WAM~\citep{zhao2026faster} & \textbf{99.6} & 99.4 & \textbf{98.2} & 96.8 & \underline{98.5} \\
            \midrule
            \textbf{Elastic-WAM (Ours)} & \underline{99.4} & 99.6 & \underline{98.0} & \underline{98.2} & \textbf{98.8} \\
            \bottomrule
        \end{tabularx}
        \endgroup
    \end{minipage}%
    \hfill
    \begin{minipage}[t]{0.485\textwidth}
        \centering
        \caption{LIBERO-Pro success rates (\%). The best and second-best results are highlighted in \textbf{bold} and \underline{underlined}, respectively.}
        \label{tab:libero-pro}
        \vspace{0.3em}
        \begingroup
        \footnotesize
        \renewcommand{\arraystretch}{1.17}
        \setlength{\tabcolsep}{1pt}
        \setlength{\arrayrulewidth}{0.35pt}
        \begin{tabularx}{\linewidth}{@{}>{\raggedright\arraybackslash}p{0.39\linewidth}*{4}{>{\centering\arraybackslash}X}|>{\centering\arraybackslash}X@{}}
            \toprule
            Method & Obj. & Pos. & Sem. & Task & Avg. \\
            \midrule
            $\pi_{0.5}$~\citep{intelligence2025pi_05} & \textbf{96.0} & \textbf{20.8} & 95.8 & 0.8 & \underline{53.4} \\
            OpenVLA~\citep{kim2024openvla} & \underline{93.0} & 0.0 & \underline{97.3} & 0.0 & 47.6 \\
            X-VLA~\citep{zheng2026x} & 79.0 & 0.8 & 96.8 & 6.8 & 45.9 \\
            Fast-WAM~\citep{yuan2026fast} & 78.0 & 1.0 & 83.2 & 15.8 & 44.5 \\
            Joint-WAM~\citep{zhao2026faster} & 84.5 & 6.3 & 96.2 & \underline{24.2} & 52.8 \\
            \midrule
            \textbf{Elastic-WAM (Ours)} & 85.6 & \underline{8.4} & \textbf{97.7} & \textbf{24.7} & \textbf{54.1} \\
            \bottomrule
        \end{tabularx}
        \endgroup
    \end{minipage}

    \par\vspace{1.4em}\noindent
    \begin{minipage}[t]{0.650\textwidth}
        \centering
        \caption{LIBERO-Plus success rates (\%). The best and second-best results are highlighted in \textbf{bold} and \underline{underlined}, respectively.}
        \label{tab:libero-plus}
        \vspace{0.3em}
        \begingroup
        \footnotesize
        \renewcommand{\arraystretch}{1.17}
        \setlength{\tabcolsep}{0.5pt}
        \setlength{\arrayrulewidth}{0.35pt}
        \begin{tabularx}{\linewidth}{@{}>{\raggedright\arraybackslash}p{0.30\linewidth}*{7}{>{\centering\arraybackslash}X}|>{\centering\arraybackslash}X@{}}
            \toprule
            Method & Cam. & Rob. & Lang. & Light & Backg. & Noise & Layout & Avg. \\
            \midrule
            \mbox{UniVLA~\citep{bu2025univla}} & 1.8 & 46.2 & 69.6 & 69.0 & 81.0 & 21.2 & 31.9 & 42.9 \\
            \mbox{OpenVLA-OFT~\citep{kim2025fine}} & \underline{56.4} & 31.9 & 79.5 & 88.7 & \textbf{93.3} & \underline{75.8} & 74.2 & \underline{69.6} \\
            \mbox{$\pi_0$~\citep{black2024pi_0}} & 13.8 & 6.0 & 58.8 & 85.0 & \underline{81.4} & \textbf{79.0} & 68.9 & 53.6 \\
            \mbox{$\pi_0$-Fast~\citep{pertsch2025fast}} & \textbf{65.1} & 21.6 & 61.0 & 73.2 & 73.2 & 74.4 & 68.8 & 61.6 \\
            \mbox{WorldVLA~\citep{cen2025worldvla}} & 0.1 & 27.9 & 41.6 & 43.7 & 17.1 & 10.9 & 38.0 & 25.0 \\
            \mbox{Fast-WAM~\citep{yuan2026fast}} & 18.8 & 45.7 & 70.1 & 83.2 & 45.7 & 29.8 & 62.7 & 49.1 \\
            \mbox{Joint-WAM~\citep{zhao2026faster}} & 37.5 & 64.5 & \underline{93.0} & \underline{95.0} & 55.9 & 47.3 & \underline{79.6} & 66.3 \\
            \mbox{LingBot-VA~\citep{li2026causal}} & 40.9 & \textbf{83.0} & 86.4 & 82.3 & 53.1 & 64.4 & 76.2 & 69.5 \\
            \midrule
            \mbox{\textbf{Elastic-WAM (Ours)}} & 52.1 & \underline{76.3} & \textbf{95.5} & \textbf{95.1} & 60.0 & 61.1 & \textbf{81.4} & \textbf{73.7} \\
            \bottomrule
        \end{tabularx}
        \endgroup
    \end{minipage}%
    \hfill
    \begin{minipage}[t]{0.325\textwidth}
        \centering
        \caption{RoboCasa-GR1 success rates (\%) and inference latency (ms). The best and second-best success rates are highlighted in \textbf{bold} and \underline{underlined}, respectively.}
        \label{tab:robocasa-gr1}
        \vspace{0.3em}
        \begingroup
        \footnotesize
        \renewcommand{\arraystretch}{1.25}
        \setlength{\tabcolsep}{1pt}
        \setlength{\arrayrulewidth}{0.35pt}
        \begin{tabularx}{\linewidth}{@{}>{\raggedright\arraybackslash}p{0.50\linewidth}>{\centering\arraybackslash}X|>{\centering\arraybackslash}X@{}}
            \toprule
            Method & \mbox{Latency $\downarrow$} & Avg. \\
            \midrule
            \mbox{$\pi_{0.5}$~\citep{intelligence2025pi_05}} & 80.5 & 37.0 \\
            \mbox{GR00T-N1.6~\citep{nvidia20256}} & 77.0 & 47.6 \\
            \mbox{StarVLA~\citep{community2026starvla}} & 82.0 & 48.8 \\
            \mbox{Fast-WAM~\citep{yuan2026fast}} & 331.1 & 54.9 \\
            \mbox{Joint-WAM~\citep{zhao2026faster}} & 572.6 & \underline{55.0} \\
            \midrule
            \textbf{Elastic-WAM (Ours)} & 388.2 & \textbf{57.4} \\
            \bottomrule
        \end{tabularx}
        \par\vspace{0.1em}
        {\fontsize{6.7}{7.8}\selectfont}
        \endgroup
    \end{minipage}
\end{table*}

During sparse joint updates, observation and action queries remain dense, while the selected future video queries retain access to the full logical key--value context.
Unselected future video tokens preserve their cached hidden states, keys, values, and last-refresh velocities, which are reused as their latents continue to evolve.
Once the cumulative \textit{World Commit} threshold is reached or a safety condition is triggered, Elastic-WAM stops world refinement and directly reuses the current video representation and layer-wise cache for subsequent action denoising.
No additional video prefill is required after \textit{World Commit}.

\section{Experiments}
\label{sec:experiments}

Our experiments address four key questions:
\begin{itemize}
    \item \textbf{\textit{Q1:}} How well does Elastic-WAM perform under increasingly challenging manipulation conditions?
    \item \textbf{\textit{Q2:}} How does our Elastic-WAM balance control performance and inference cost?
    \item \textbf{\textit{Q3:}} How does adaptive temporal and spatial refinement contribute to control performance?
    \item \textbf{\textit{Q4:}} How robust is our Elastic-WAM in real-world long-horizon manipulation under lighting and object changes?
\end{itemize}

\subsection{Experimental Setup}
\label{sec:experimental-setup}

We evaluate Elastic-WAM on LIBERO~\citep{liu2023libero} for standard manipulation, LIBERO-Plus~\citep{fei25libero-plus} for robustness under visual and environmental perturbations, and LIBERO-Pro~\citep{zhou2025libero} for generalization across object, position, language, and task variations.
We further evaluate complex manipulation performance and inference efficiency on RoboCasa-GR1~\citep{bjorck2025gr00t}.
For real-world evaluation, we consider a four-stage bimanual tea preparation task under nominal conditions, lighting changes, and object replacement.

We compare Elastic-WAM against existing policies, including Joint-WAM~\citep{zhao2026faster} and Fast-WAM~\citep{yuan2026fast}, while keeping the pretrained WAM frozen during refinement field learning.
On LIBERO-Plus and LIBERO-Pro, the LIBERO-trained policy is evaluated without additional adaptation.
We report task success rates across simulation benchmarks, inference latency on RoboCasa-GR1, and the full-task success rate (SR) and the average task progress (PS) in real-world experiments.

\subsection{Comparison with State-of-the-Art Methods}
\label{sec:sota-comparison}
In this section, we address \textbf{\textit{Q1}} and \textbf{\textit{Q2}} by evaluating control performance across four simulation benchmarks and inference efficiency on the most challenging benchmark, RoboCasa-GR1.

\noindent\textbf{LIBERO.}
Elastic-WAM achieves the highest average success rate on LIBERO (Table~\ref{tab:libero}), with notable improvements over Joint-WAM on the Long suite.
These results demonstrate that adaptive refinement can maintain strong control performance without uniformly refining the future.

\noindent\textbf{LIBERO-Pro and LIBERO-Plus.}
On LIBERO-Pro (Table~\ref{tab:libero-pro}), Elastic-WAM achieves the highest average success rate, consistently outperforming Joint-WAM across object, position, semantic, and task variations.
On LIBERO-Plus (Table~\ref{tab:libero-plus}), Elastic-WAM demonstrates superior robustness across all perturbation categories, with particularly notable gains under camera viewpoint changes, robot initialization changes, and sensor noise.
These improvements suggest that adapting refinement depth and spatial allocation to the current state provides more effective future representations for action generation under diverse distribution shifts.

\noindent\textbf{RoboCasa-GR1.}
Elastic-WAM achieves the highest success rate on RoboCasa-GR1 (Table~\ref{tab:robocasa-gr1}), while reducing inference latency by 32.2\% compared with Joint-WAM.
These results highlight the benefit of selectively refining action-relevant future representations, improving control in complex manipulation while avoiding redundant world updates.

\subsection{Effectiveness of Adaptive Refinement}
\label{sec:adaptive-refinement}
To address \textbf{\textit{Q3}}, we investigate the contributions of spatial selection, adaptive temporal control, and token-refresh budget through ablation studies, complemented by a qualitative analysis of refinement behavior.

\begin{table*}[t]
    \centering
    \begin{minipage}[t]{1.0\textwidth}
        \centering
        \caption{Spatial selection and future refinement on RoboCasa-GR1 (success rates, \%). The best and second-best results are highlighted in \textbf{bold} and \underline{underlined}, respectively.}
        \label{tab:ablation-robocasa}
        \vspace{0.3em}
        \begingroup
        \footnotesize
        \renewcommand{\arraystretch}{1.17}
        \setlength{\tabcolsep}{0.7pt}
        \setlength{\arrayrulewidth}{0.35pt}
        \begin{tabularx}{\linewidth}{@{}>{\raggedright\arraybackslash}p{0.27\linewidth}*{7}{>{\centering\arraybackslash}X}|>{\centering\arraybackslash}X@{}}
            \toprule
            Variant & Cab. & Drawer & Micro. & CutBoard & Placemat & Plate & Tray & Overall \\
            \midrule
            \textbf{Learned WHERE, $K=96$} & \textbf{65.00} & \textbf{65.00} & \underline{50.00} & \textbf{63.60} & \underline{48.50} & \textbf{60.00} & \textbf{53.20} & \textbf{57.42} \\
            Random WHERE, $K=96$ & 61.00 & \underline{63.00} & \underline{50.00} & 55.60 & 47.50 & 59.00 & 48.00 & 53.83 \\
            All-active, $K=128$ & 61.00 & 60.00 & 45.00 & \underline{60.40} & \textbf{50.00} & \underline{59.50} & 50.40 & \underline{55.17} \\
            FirstKV & \underline{62.00} & 54.00 & \textbf{53.00} & 59.60 & 46.00 & 58.50 & \underline{50.80} & 54.50 \\
            \bottomrule
        \end{tabularx}
        \endgroup
        \par\vspace{0.25em}
        {\footnotesize\raggedright\textsuperscript{*} The first three variants use adaptive temporal control. $K$ counts refreshed future tokens after the initial dense joint denoising update. FirstKV follows the video update strategy of Faster-WAM~\citep{zhao2026faster}, reusing the initial video KV cache without further refinement.\par}
    \end{minipage}

    \par\vspace{1.3em}\noindent
    \begin{minipage}[t]{1.0\textwidth}
        \centering
        \caption{Temporal control and token-refresh budget on LIBERO-Plus (success rates, \%). The best and second-best results within each panel are highlighted in \textbf{bold} and \underline{underlined}, respectively.}
        \label{tab:ablation-libero}
        \vspace{0.3em}
        \begingroup
        \footnotesize 
        \renewcommand{\arraystretch}{1.17}
        \setlength{\tabcolsep}{0.7pt}
        \setlength{\arrayrulewidth}{0.35pt}
        \begin{tabularx}{\linewidth}{@{}>{\raggedright\arraybackslash}p{0.27\linewidth}*{7}{>{\centering\arraybackslash}X}|>{\centering\arraybackslash}X@{}}
            \toprule
            Variant & Cam. & Rob. & Lang. & Light & Backg. & Noise & Layout & Overall \\
            \midrule
            \multicolumn{9}{@{}l}{\textit{(a) Temporal control: learned WHERE, $K=96$}} \\
            \addlinespace[2pt]
            WHERE-only & \underline{47.92} & 71.72 & \underline{94.67} & 93.21 & 59.46 & 58.75 & \textbf{81.82} & 71.71 \\
            Fixed step & 47.50 & \underline{75.25} & 94.26 & \underline{94.44} & \textbf{61.08} & \underline{59.14} & 80.91 & \underline{72.31} \\
            \textbf{Adaptive} & \textbf{52.08} & \textbf{76.26} & \textbf{95.49} & \textbf{95.06} & \underline{60.00} & \textbf{61.09} & \underline{81.36} & \textbf{73.71} \\
            \midrule
            \multicolumn{9}{@{}l}{\textit{(b) Token-refresh budget: adaptive temporal control}} \\
            \addlinespace[2pt]
            $K=49$ & \underline{48.75} & \underline{73.23} & 94.26 & \textbf{96.30} & \textbf{62.16} & 59.14 & \textbf{82.73} & \underline{72.84} \\
            \textbf{$K=96$} & \textbf{52.08} & \textbf{76.26} & \textbf{95.49} & \underline{95.06} & 60.00 & \textbf{61.09} & \underline{81.36} & \textbf{73.71} \\
            $K=147$ & 46.67 & 68.69 & \underline{94.67} & 94.44 & \underline{60.54} & \underline{59.92} & \textbf{82.73} & 71.71 \\
            $K=196$ & 46.25 & 69.19 & 94.26 & \underline{95.06} & 59.46 & 57.98 & 80.45 & 70.92 \\
            \bottomrule
        \end{tabularx}
        \endgroup
        \par\vspace{0.25em}
        {\footnotesize\raggedright\textsuperscript{*} In (a), Fixed step uses adaptive commit with fixed progression; Adaptive uses both adaptive commit and progression. In (b), $K$ counts refreshed future tokens after the initial dense joint denoising update. The Adaptive and $K=96$ rows share the same configuration.\par}
    \end{minipage}
\end{table*}

\noindent\textbf{Spatial Refinement.}
On RoboCasa-GR1 (Table~\ref{tab:ablation-robocasa}), learned spatial selection outperforms random selection under the same token budget, as well as all-active refinement with a larger budget and FirstKV, which reuses the initial video cache.
These results demonstrate the benefit of selectively refreshing action-relevant future regions rather than relying on random allocation, broader refinement, or cached representations alone.

\noindent\textbf{Temporal Refinement.}
On LIBERO-Plus (Table~\ref{tab:ablation-libero}(a)), we progressively introduce adaptive World Commit (WHEN) and denoising progression (HOW FAR), while keeping learned spatial selection and the token budget fixed.
Adaptive stopping improves upon the WHERE-only baseline, while further adapting denoising progression yields the best overall performance, demonstrating the complementary benefits of temporal adaptation.

\begin{figure}[!htb]
    \centering
    \includegraphics[width=1.0\linewidth]{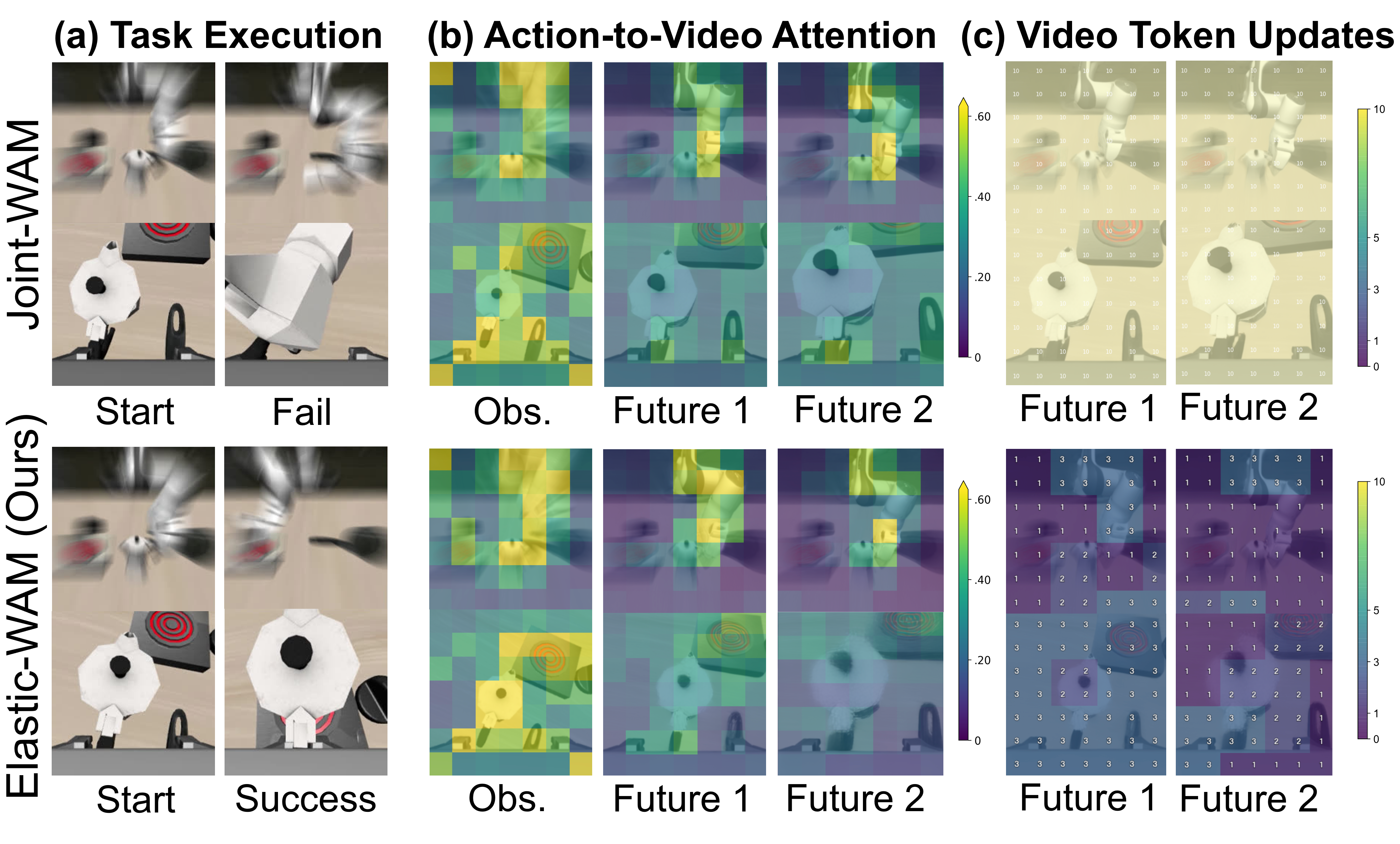}
    \caption{\textbf{Qualitative comparison} between Joint-WAM (top) and Elastic-WAM (bottom). (a) Task outcomes. (b) Action-to-video attention maps. (c) Spatial distribution of future video token update counts. Elastic-WAM achieves successful control through selective refinement while preserving the full attention context.}
    \label{fig:compare-jointwam}
\end{figure}

\noindent\textbf{Refinement Budget.}
We further examine the effect of different token-refresh budgets on LIBERO-Plus (Table~\ref{tab:ablation-libero}(b)), while keeping adaptive temporal control fixed.
The default budget $K=96$ achieves the highest overall success rate, with smaller budgets remaining competitive and larger budgets yielding lower performance.
These results indicate that increasing the number of refreshed future video tokens does not necessarily improve downstream control.

\noindent\textbf{Qualitative Analysis.}
Figure~\ref{fig:compare-jointwam} compares Elastic-WAM and Joint-WAM under noisy observations on LIBERO-Plus.
Although both methods exhibit similar action-to-video attention patterns, Elastic-WAM selectively refreshes future video regions while retaining cached representations for others.
This allows different regions to undergo different levels of refinement while preserving the full logical attention context for action generation.
In the illustrated case, Elastic-WAM succeeds where Joint-WAM fails, suggesting that selective refinement can preserve useful action information while avoiding redundant world updates.

\subsection{Real-World Long-Horizon Manipulation}
\label{sec:real-world}

\begin{figure}[t]
    \centering
    \includegraphics[width=\linewidth]{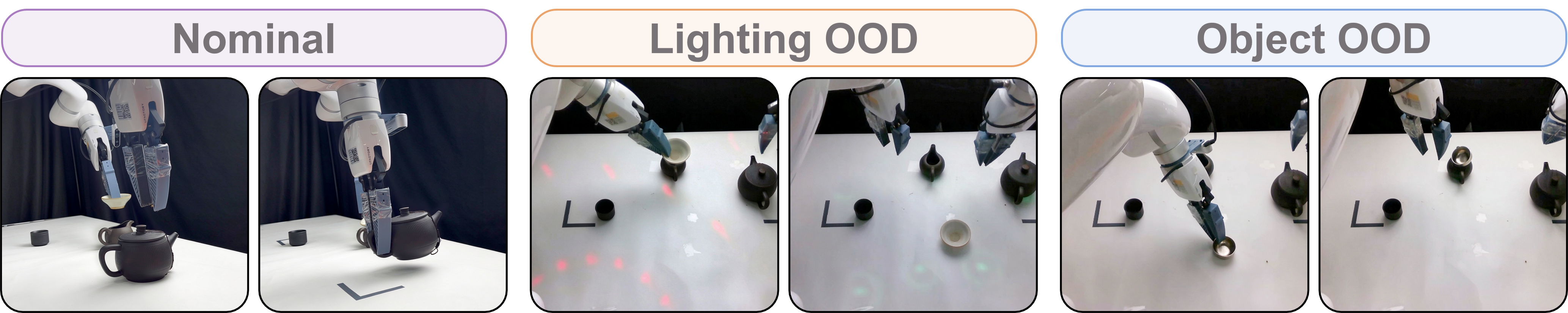}
    \caption{
        \textbf{Real-world evaluation settings.}
        We evaluate under three conditions:
        Nominal, Lighting OOD, and Object OOD.
        Lighting OOD introduces strong illumination changes, while Object OOD replaces the white cup with a silver-colored cup of a slightly different shape.
    }
    \label{fig:real}
\end{figure}

To address \textbf{\textit{Q4}}, we conduct real-world experiments on a bimanual robotic platform equipped with two UFACTORY xArm 7 manipulators.



\noindent\textbf{Setup.}
We evaluate a four-stage tea preparation task involving cup placement, tea pouring, and object return.
As shown in Fig.~\ref{fig:real}, we consider three evaluation conditions: nominal, lighting Out-of-Distribution (OOD) with neon illumination, and object OOD with cup replacement.
We compare Elastic-WAM against Joint-WAM and Fast-WAM, evaluating each method over 20 trials per condition.
We report full-task success rate (SR) and progress score (PS), measuring complete task success and the average fraction of completed stages, respectively.
Additional experimental details are provided in Appendix~\ref{app:real-world}.

\begin{table}[t]
    \centering
    \caption{Real-world long-horizon manipulation results. SR denotes full-task success rate (\%), and PS denotes average task progress (\%). The best results are highlighted in \textbf{bold}.}
    \label{tab:realworld}
    \footnotesize
    \renewcommand{\arraystretch}{1.15}
    \begin{tabular*}{\linewidth}{@{\extracolsep{\fill}}lcccccc@{}}
        \toprule
        & \multicolumn{2}{c}{Nominal}
        & \multicolumn{2}{c}{Lighting OOD}
        & \multicolumn{2}{c}{Object OOD} \\
        \cmidrule(lr){2-3}
        \cmidrule(lr){4-5}
        \cmidrule(l){6-7}
        Method & SR & PS & SR & PS & SR & PS \\
        \midrule
        Fast-WAM~\citep{yuan2026fast}
        & 0.0 & 12.5
        & 0.0 & 5.0
        & 0.0 & 10.0 \\
        Joint-WAM~\citep{zhao2026faster}
        & 30.0 & 37.5
        & 30.0 & 35.0
        & \textbf{30.0} & 35.0 \\
        \midrule
        \textbf{Elastic-WAM (Ours)}
        & \textbf{50.0} & \textbf{57.5}
        & \textbf{50.0} & \textbf{52.5}
        & \textbf{30.0} & \textbf{47.5} \\
        \bottomrule
    \end{tabular*}
\end{table}

\noindent\textbf{Results.}
As shown in Table~\ref{tab:realworld}, Elastic-WAM achieves the highest full-task SR, while consistently attaining the best task PS across all evaluation settings.
Under object OOD, although Elastic-WAM and Joint-WAM achieve comparable full-task success, Elastic-WAM completes more manipulation stages on average, revealing improvements that are not fully captured by the full-task success metric.
In contrast, Fast-WAM fails to complete the entire task under any condition.
These results demonstrate that Elastic-WAM enables more reliable long-horizon execution, even under challenging visual distribution shifts.

\noindent\textbf{Long-Horizon Analysis.}
Long-horizon manipulation requires policies to recognize task progress from evolving visual observations and generate actions appropriate to each execution stage.
Without an explicit task-progress memory module, the evaluated policies must largely infer their current stage from visual information.
We hypothesize that jointly adapting the temporal progression and spatial allocation of future refinement helps Elastic-WAM capture stage-dependent visual changes more effectively, thereby facilitating task-state discrimination and stage-appropriate action generation.
This interpretation is consistent with its superior performance on the LIBERO Long suite (Table~\ref{tab:libero}) and higher task progress in real-world experiments (Table~\ref{tab:realworld}), suggesting that adaptive spatiotemporal refinement may also benefit long-horizon task execution.
\section{Conclusion}
\label{sec:conclusion}

In this paper, we presented Elastic-WAM, a unified framework for adaptively allocating world refinement computation across temporal denoising time and spatial future video space.
By modeling the remaining refinement demand through a spatiotemporal field, Elastic-WAM jointly determines when to stop world refinement, how far to advance each denoising step, and which future video regions to refresh.
We learn this field through three stages of action-grounded supervision and enable efficient inference through recurrent sparse refinement while preserving the full logical attention context.
Extensive experiments across diverse simulation benchmarks and real-world long-horizon manipulation demonstrate state-of-the-art control performance with reduced inference latency.
{
    \small
    \bibliographystyle{ieeenat_fullname}
    \bibliography{main}
}


\clearpage

\twocolumn[{
    \begin{center}
        {\Large\bfseries Appendix of Elastic-WAM
        }
    \end{center}
    \vspace{1em}
}]

\appendix

\section{Implementation Details}
\label{app:implementation}

\subsection{Base World-Action Model}
\label{app:base-model}

We build Elastic-WAM upon FastWAM-Joint, a pretrained world-action model that jointly refines future video representations and robot actions through iterative denoising.
FastWAM-Joint is the same model referred to as Joint-WAM in Faster-WAM~\citep{zhao2026faster}; we use the former name to describe our backbone and the latter for baseline comparisons.
The model follows the joint video-action denoising configuration used by the Joint-WAM baseline~\citep{zhao2026faster}.
As illustrated in Fig.~\ref{fig:pipeline}, language instructions are processed by a T5 encoder, visual observations are encoded through a VAE, and proprioceptive states are mapped into the model's feature space through a projection layer.
The joint denoising backbone processes observation, future-video, and action tokens, allowing future visual representations to provide conditioning for action generation.
In our configuration, the future-video stream contains 196 tokens, which are organized into 64 geometry-aware groups for adaptive spatial refinement.
Throughout the three-stage refinement field training, all pretrained WAM parameters remain frozen.
Elastic-WAM preserves the underlying action generation mechanism while adaptively controlling the temporal and spatial computation devoted to future video refinement.

\subsection{Simulation Benchmarks}
\label{app:benchmarks}

We evaluate Elastic-WAM across four simulation benchmarks covering standard manipulation, out-of-distribution generalization, and complex robotic interactions.

\noindent\textbf{LIBERO.}
LIBERO~\citep{liu2023libero} comprises four manipulation suites: Spatial, Object, Goal, and Long.
Each suite contains 10 language-conditioned tasks, covering spatial reasoning, object manipulation, goal-conditioned execution, and extended action sequences.
We evaluate each task over 50 rollouts, resulting in 2,000 evaluation episodes across all four suites.
We report the success rate of each suite and the average performance across suites.

\noindent\textbf{LIBERO-Plus.}
LIBERO-Plus~\citep{fei25libero-plus} evaluates robustness under seven categories of distribution shifts: camera viewpoint, robot initialization, language instruction, illumination, background appearance, sensor noise, and object layout.
We directly evaluate the LIBERO-trained policy without additional adaptation to these perturbations.
We report success rates for individual perturbation categories and aggregate overall performance across evaluation episodes, accounting for the number of episodes in each category.

\noindent\textbf{LIBERO-Pro.}
LIBERO-Pro~\citep{zhou2025libero} evaluates generalization across four categories: Object, Position, Semantic, and Task.
These settings examine robustness to changes in manipulated objects, object positions, language semantics, and task configurations.
As with LIBERO-Plus, we evaluate the LIBERO-trained policy without additional fine-tuning.
We report category-wise success rates and their unweighted average.

\noindent\textbf{RoboCasa-GR1.}
RoboCasa-GR1~\citep{bjorck2025gr00t} evaluates complex tabletop manipulation across 24 tasks involving object rearrangement and interactions with articulated objects.
The benchmark covers diverse manipulation scenarios, including cabinets, drawers, microwaves, cutting boards, placemats, plates, and trays.
We report task success rates and overall performance to assess the effectiveness of adaptive refinement in challenging manipulation settings.
We follow the evaluation setting used by StarVLA~\citep{community2026starvla}.

\subsection{Inference Efficiency and Hardware}
\label{app:hardware}

We measure inference latency on an NVIDIA H800 GPU using the RoboCasa-GR1 benchmark.
We report inference latency in milliseconds alongside task success rates to evaluate the trade-off between control performance and computational efficiency.
As reported in Table~\ref{tab:robocasa-gr1}, Elastic-WAM reduces inference latency by 32.2\% compared with Joint-WAM while achieving higher task success.
This efficiency improvement is enabled by adaptive temporal refinement, selective future-video token updates, and cache reuse during action generation.

\section{Additional Simulation Experiments}
\label{app:simulation}

\subsection{Precision-Sensitive Manipulation}
\label{app:precision}

\begin{figure*}[t]
    \centering
    \includegraphics[width=0.9\textwidth]{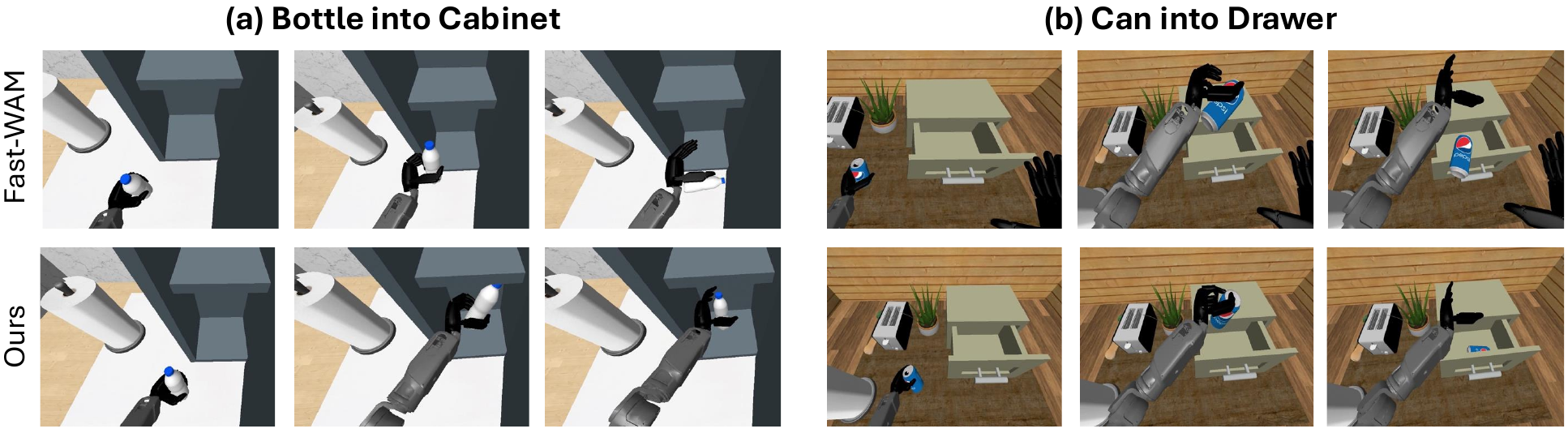}
    \caption{
        \textbf{Precision-sensitive manipulation on RoboCasa-GR1.}
        Qualitative comparisons between Fast-WAM (top) and Elastic-WAM (bottom) on (a) Bottle into Cabinet and (b) Can into Drawer.
        Keyframes progress from left to right, highlighting the final placement stage.
        Fast-WAM fails to complete the placement, whereas Elastic-WAM successfully finishes both tasks.
    }
    \label{fig:precision}
\end{figure*}

We provide additional qualitative comparisons on the RoboCasa-GR1 to examine manipulation behaviors requiring precise object placement.
As shown in Fig.~\ref{fig:precision}, we compare Elastic-WAM with Fast-WAM on two representative tasks: placing a bottle into a cabinet and placing a can into a drawer.

In both examples, Fast-WAM moves the objects toward their target locations but fails to complete the final placement, whereas Elastic-WAM successfully executes the manipulation.
These observations suggest that selectively continuing future refinement can provide more effective conditioning for precision-sensitive actions, where reaching the target vicinity alone is insufficient for successful task completion.

\section{Additional Real-World Experiments}
\label{app:real-world}

\subsection{Robotic Platform and Data Collection}
\label{app:robot-platform}

We conduct real-world experiments on a bimanual robotic platform equipped with two UFACTORY xArm 7 manipulators.
The tabletop workspace contains two teapots and multiple cups, requiring coordinated manipulation of different objects throughout a long-horizon tea-preparation task.
The evaluation involves successive grasping, placement, pouring, and object-return operations, with changes in object configurations across execution stages.
We additionally record head-camera and wrist-camera views to visualize the robot's execution behavior.

We collect 200 demonstration trajectories for the four-stage tea-preparation task.
The underlying world-action policies are prepared following the standard FastWAM-Joint training procedure.
For Elastic-WAM, the pretrained WAM backbone remains frozen during subsequent refinement field learning, as described in Sec.~\ref{sec:method}.
This setup allows us to evaluate the effect of adaptive world refinement on long-horizon manipulation without modifying the underlying action generation architecture.

\subsection{Task and Evaluation Protocol}
\label{app:real-protocol}

\noindent\textbf{Long-Horizon Task.}
We design a four-stage tea-preparation task requiring the robot to execute a sequence of coordinated bimanual operations.
The complete language instruction is:
\emph{``Place the white cup on top of the uncovered teapot, pour tea from the covered teapot into the white cup, return the white cup to its original position, then pour tea from the uncovered teapot into the black cup and return the teapot to its original position.''}

The task consists of four sequential stages.
In the first stage, the robot grasps the white cup and places it on the uncovered teapot.
In the second stage, it manipulates the covered teapot to pour tea into the white cup.
In the third stage, the robot retrieves the white cup and returns it to its original position.
Finally, it pours tea from the uncovered teapot into the black cup and returns the teapot to its original position.
Successful task execution requires completing all four stages in sequence without human intervention.

\noindent\textbf{Evaluation Conditions.}
We evaluate Elastic-WAM, Joint-WAM, and Fast-WAM under three conditions: Nominal, Lighting OOD, and Object OOD, as illustrated in Fig.~\ref{fig:real}.
The Nominal condition uses the standard tabletop environment and object configuration.
Lighting OOD introduces neon illumination to alter the visual appearance of the scene.
Object OOD replaces the original white cup with a silver-colored cup of a slightly different shape throughout the manipulation sequence.
These perturbations evaluate the robustness of long-horizon execution to changes in scene appearance and object characteristics.

Each method is evaluated over 20 trials per condition.
All methods are required to complete the same four-stage task under the corresponding evaluation setting.

\noindent\textbf{Evaluation Metrics.}
We report two complementary metrics: full-task success rate (SR) and progress score (PS).
A trial is considered successful only when all four stages are completed.
The full-task success rate is defined as the percentage of trials satisfying this criterion.

The progress score measures the average fraction of successfully completed stages across evaluation trials.
Each stage contributes one point only when all operations within that stage are completed.
For a four-stage task, the number of completed stages is divided by four and averaged across trials, with the resulting score expressed as a percentage.
Unlike full-task success, PS captures partial task completion and provides a more detailed measure of execution progress when a policy fails before completing the entire sequence.
The quantitative results are presented in Table~\ref{tab:realworld}.

\subsection{Qualitative Long-Horizon Execution}
\label{app:real-qualitative}

\begin{figure*}[t]
    \centering
    \includegraphics[width=1.0\textwidth]{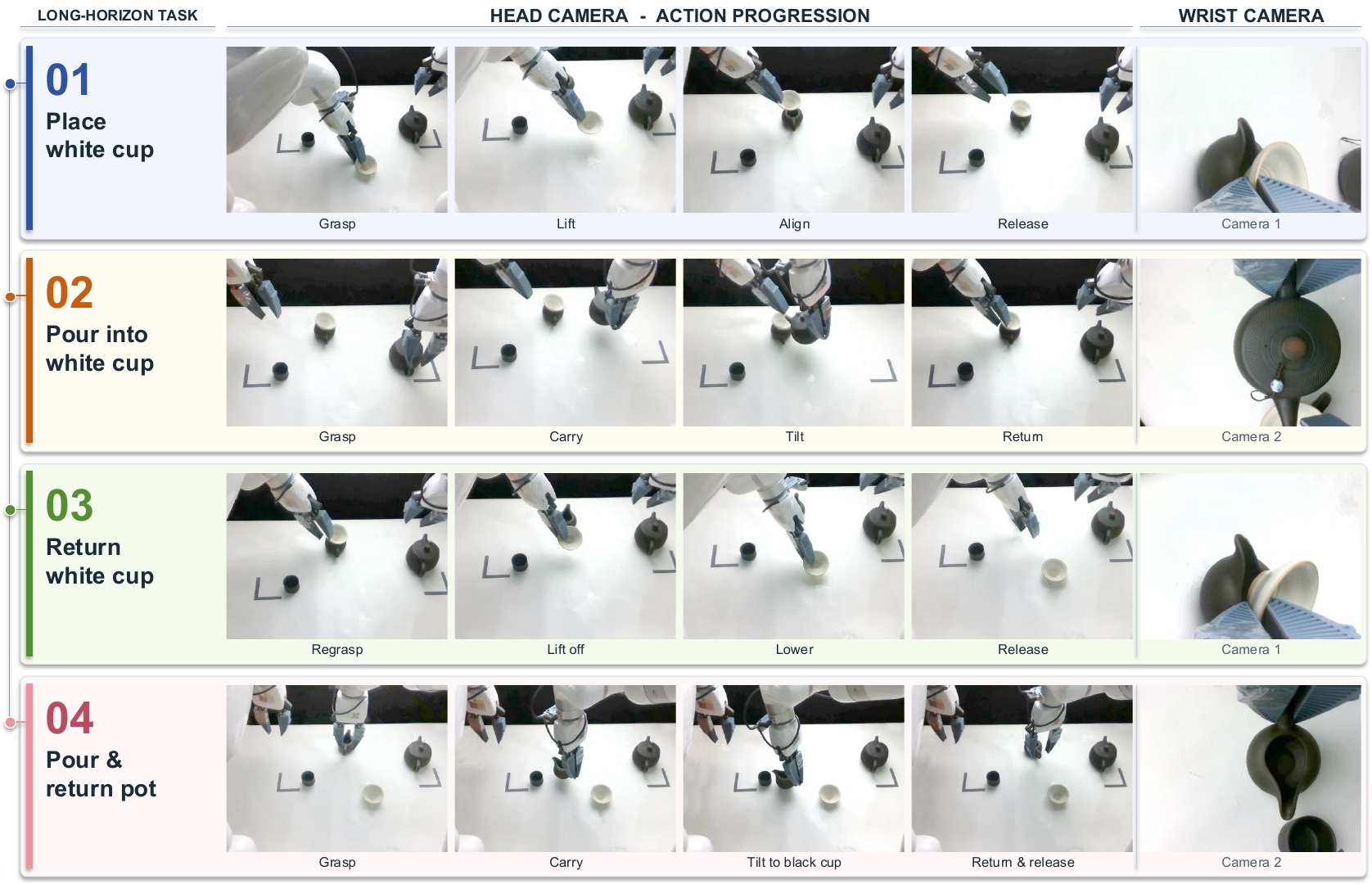}
    \caption{
        \textbf{Qualitative visualization of real-world long-horizon manipulation.}
        Keyframes illustrate a successful four-stage tea-preparation rollout using Elastic-WAM.
        \textbf{Stage 1:} Grasping, lifting, aligning, and placing the white cup on the uncovered teapot.
        \textbf{Stage 2:} Grasping and manipulating the covered teapot to pour tea into the white cup.
        \textbf{Stage 3:} Retrieving and returning the white cup to its original position.
        \textbf{Stage 4:} Pouring tea into the black cup and returning the uncovered teapot.
        The head-camera keyframes and corresponding wrist-camera views illustrate the progression of robot actions throughout the task.
    }
    \label{fig:real-rollout}
\end{figure*}

We provide additional qualitative results illustrating a successful long-horizon tea-preparation rollout with Elastic-WAM.
Figure~\ref{fig:real-rollout} presents representative keyframes from all four execution stages, including the corresponding wrist-camera observations.
The visualization captures the progression of the manipulation sequence from initial cup placement to the final pouring and object-return operations.

Throughout the rollout, the robot successfully performs successive grasping, placement, pouring, and object-return operations as the scene configuration evolves.
Each completed stage changes the positions or states of task-relevant objects, requiring subsequent actions to be generated according to the updated visual observations.
This example illustrates Elastic-WAM's ability to sustain coordinated manipulation across multiple execution stages and complements the quantitative long-horizon results reported in Sec.~\ref{sec:real-world}.

\end{document}